%% file: main.tex
\documentclass[10pt,letterpaper]{article}

\usepackage[preprint]{tmlr}
\makeatletter
\let\ifdeanonymized\if@accepted % true with [accepted] or [preprint]; controls the acknowledgements
\makeatother

\input{math_commands.tex}

\usepackage[T1]{fontenc}
\usepackage[utf8]{inputenc}
\usepackage{lmodern}
\usepackage{microtype}
\usepackage{graphicx}
\usepackage{pgfplots}
\pgfplotsset{compat=1.18}
\usepgfplotslibrary{groupplots}
\usetikzlibrary{patterns}
\usepackage{float}
\usepackage{booktabs}
\usepackage{amsmath}
\usepackage{amssymb}
\usepackage{algorithm}
\usepackage{algpseudocode}
\usepackage[font=small,labelfont=bf]{caption}
\usepackage{enumitem}
\usepackage{listings}
\usepackage[hidelinks]{hyperref}

\title{\textbf{ODSA: Orchestrated Discovery and Self-Aggregation}}
\author{\name Macardle Meng \email macardle.m@columbia.edu \\
      \addr Columbia Engineering\\
      Columbia University
      \AND
      \name Haomin Zhang \email haominz@kth.se \\
      \addr School of Electrical Engineering and Computer Science\\
      KTH Royal Institute of Technology}
\date{}

\begin{document}
\maketitle

\begin{abstract}
\noindent Test-time scaling improves the accuracy of a language model's answers by spending more computation at inference time. Established methods do so either \emph{sequentially}, revising a single solution over several passes, or \emph{in parallel}, generating several independent solutions and selecting among them. We present ODSA (Orchestrated Discovery and Self-Aggregation), an LLM-directed scaling framework in which an orchestrator LLM is given existing test-time scaling methods as actions --- \emph{discovery} of a new independent solution, \emph{comparison} of existing solutions to build a better one, and \emph{acceptance} of a proposed answer, which triggers its verification --- and directs stateless workers to carry them out. The orchestrator does not solve the problem itself: it allocates computation, feeds the results of earlier actions into later ones, and takes every action from a single context, so that each decision is informed by everything tried before. On Humanity's Last Exam, ODSA improves accuracy by 4.7 to 18.0 points across three models over a single direct call to the model with no test-time scaling. On LiveCodeBench, where a single direct call already solves most problems, it improves accuracy by 0.6 to 14.0 points.
\end{abstract}

% ============================================================================
\section{Introduction}
\label{sec:intro}

Test-time scaling improves the accuracy of a language model's answers by spending more computation at inference time instead of retraining the model \citep{snell2025}. \emph{Sequential} scaling revises a single solution over several passes of self-critique \citep{madaan2023}, but a revised chain never sees an alternative to itself. \emph{Parallel} scaling samples many independent solutions and selects among them \citep{wang2023} or aggregates them into a new one with Generative Self-Aggregation \citep[GSA;][]{li2025}, but a correct sample is often generated and not selected \citep{brown2024}. Recursive Self-Aggregation \citep[RSA;][]{venkatraman2025} combines the two by aggregating groups of solutions into new ones, round after round. That makes its cost grow with the number of solutions and the number of rounds, and since the schedule is fixed in advance, every problem gets the same computation whether it needs it or not.

Adaptive methods cut this cost by deciding how much to spend as they go: they stop sampling once the answers agree \citep{aggarwal2023,li2024}, or stop refining once the model says it is confident \citep{jin2026}. More recently, LLM-directed methods went a step further and put an LLM in charge of the scaling process itself \citep{qin2026,wong2026}.

We present ODSA (Orchestrated Discovery and Self-Aggregation), an LLM-directed framework that combines parallel and sequential scaling under one orchestrator LLM. The orchestrator is given three actions: \textbf{discovery} samples a new independent solution along a direction the orchestrator names; \textbf{comparison} aggregates several existing solutions into a better one; and \textbf{acceptance} proposes a candidate solution as final and triggers a \emph{verification} pass that critiques it. The orchestrator never attempts the problem: it reads the result of each action, stored as a candidate solution, decides which action to take next and on which candidate solutions, and takes every action from one running context.

Our contributions:

\begin{enumerate}[leftmargin=1.6em,itemsep=0.35em]
\item \textbf{Allocation, not solving.} The orchestrator's only role is to allocate computation. Every action's result is a candidate solution that later actions reuse: discovered solutions are compared, comparison outputs are verified, and the verifier's rival to a disputed proposal becomes a new candidate solution.
\item \textbf{Test-time scaling methods as orchestrator actions.} Discovery, comparison, and acceptance package parallel sampling, GSA aggregation, and sequential revision as actions, so that one LLM directs both parallel and sequential scaling and applies each where the evidence calls for it.
\item \textbf{Continuity of allocation.} Every action is issued from a single context, so each decision is conditioned on all previous ones --- which directions were tried, which candidate solutions conflict, and which proposals were disputed.
\end{enumerate}

On Humanity's Last Exam \citep[HLE;][]{phan2025}, ODSA improves accuracy by 4.7 to 18.0 points across three models over a single direct call to the model with no test-time scaling (the \emph{single-sample baseline}); on LiveCodeBench \citep{jain2025}, where the single-sample baseline already solves most problems, it adds 0.6 to 14.0 points.

% ============================================================================
\section{Related work}
\label{sec:related}

\paragraph{Sequential scaling.} Sequential methods spend additional computation by revising a single solution over several passes. Self-Refine \citep{madaan2023} has the model critique its own answer and feeds that critique into the next pass. Budget forcing \citep{muennighoff2025} instead keeps a single chain thinking until it has used up a token budget. Other variants keep a reflective memory across attempts \citep{shinn2023} or add a self-verification pass after the fact \citep{weng2023}. Two results temper expectations for this family: \citet{huang2024} find that self-correction without external feedback often makes reasoning worse, and \citet{zheng2023} show that LLM judges are biased by position and verbosity.

\paragraph{Parallel scaling.} Parallel methods generate several independent solutions and select among them. Self-consistency \citep{wang2023} takes the majority vote over the final answers of chain-of-thought samples \citep{wei2022}; verifier-based selection instead ranks samples with a trained reward model \citep{cobbe2021,lightman2024}. \citet{brown2024} show that the chance that \emph{some} sample is correct grows far faster with the number of samples than the accuracy of any selection rule, locating the bottleneck in selection. GSA \citep{li2025} avoids selection altogether: it prompts the model with several of its own sampled responses and has it generate a new one. ODSA's comparison action is this aggregation applied to candidate solutions the orchestrator selects. Other synthesis-based variants, and methods that diversify samples before they are generated, follow the same pattern \citep{wang2025c,jiang2023,du2024,naik2023,wang2025a}.

\paragraph{Hybrid scaling.} Hybrid methods combine sequential and parallel computation. Tree of Thoughts \citep{yao2023} and its graph extension \citep{besta2024} search over partial thoughts in both breadth and depth under a heuristic evaluator. RSA \citep{venkatraman2025} runs GSA-style aggregation over and over. It keeps a population of solutions and, in each round, replaces every member with an aggregation of a random subset of the population, so the cost is the population size multiplied by the number of rounds.

\paragraph{Adaptive allocation.} Adaptive methods decide at run time how much computation to spend. \citet{snell2025} split a budget between sequential and parallel computation by estimated difficulty; stopping rules end sampling once the running vote is confident enough \citep{aggarwal2023,li2024,lee2025} or end refinement once the model reports confidence \citep{jin2026}. Each adapts a scalar from a summary statistic.

\paragraph{LLM-directed test-time scaling.} These methods place an LLM in charge of a component of the scaling process. In ATLAS \citep{qin2026}, an orchestrator LLM dispatches fresh solvers, decides when to stop, and synthesizes the final answer itself. In Team of Thoughts \citep{wong2026}, an orchestrator LLM picks which of several profiled models to call, so each attempt starts from a deliberately chosen region of the solution space instead of a random draw. ODSA's discovery action does the same thing within a single model: the orchestrator tells each worker which direction to take. Multi-agent frameworks such as AutoGen \citep{wu2024} and MetaGPT \citep{hong2024} split a task into sub-problems and give each LLM instance its own piece to work on. ODSA never splits the task. Its orchestrator does not work on the problem at all; it only directs the workers that do. Other work uses an LLM to design the strategy for each attempt on a fixed schedule, or to search for or discover allocation policies offline \citep{zhuang2026,wang2025b,zheng2026}.

% ============================================================================
\section{Method}
\label{sec:method}

\subsection{Orchestrator and workers}
\label{sec:roles}

\begin{figure}[!htb]
\centering
\includegraphics[width=\linewidth]{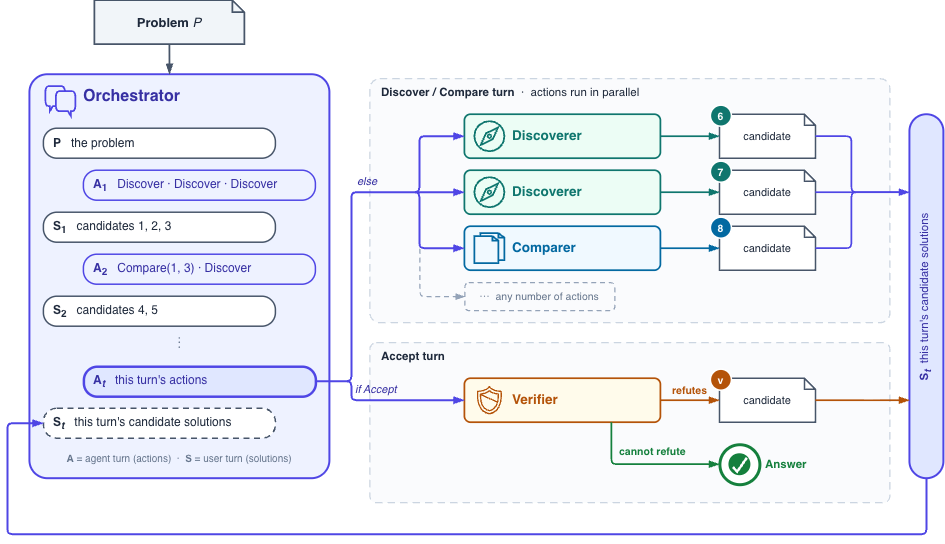}
\caption{Architecture of ODSA. Each worker's output becomes a numbered candidate solution in the orchestrator's conversation, where later actions can reference it; comparers read the candidate solutions their action names, and the verifier sees only the problem and the proposed solution.}
\label{fig:architecture}
\end{figure}

ODSA solves a problem by having one LLM direct others rather than solve it. An \textbf{orchestrator} holds a single running conversation about the problem. It does not attempt the problem itself; it issues \textbf{actions}, and actions spawn \textbf{workers} --- fresh calls to the same model --- that generate solutions. The investigation ends when the orchestrator accepts one of those solutions as the answer (Figure~\ref{fig:architecture}).

The two roles divide the work cleanly. The orchestrator is the one call with memory, and its conversation is turn-based: the first message is the problem, the orchestrator replies with actions, the workers' solutions return as the next message, the orchestrator replies with its next actions, and so on until it accepts (Algorithm~\ref{alg:odsa}). Every turn is thus decided with the whole investigation in view, and the orchestrator's only decisions are what to do next and when to stop; because it never solves, the answer ODSA returns is always one a worker wrote. Workers are stateless: each sees only the problem and the action it was handed, and it returns a new \textbf{candidate solution}.

\subsection{Three actions}
\label{sec:actions}

The orchestrator's entire vocabulary is three actions (Figure~\ref{fig:actions}; prompts in Appendix~\ref{app:prompts}). Each action commissions one worker, which produces one candidate solution.

\begin{figure}[!htb]
\centering
\includegraphics[width=\linewidth]{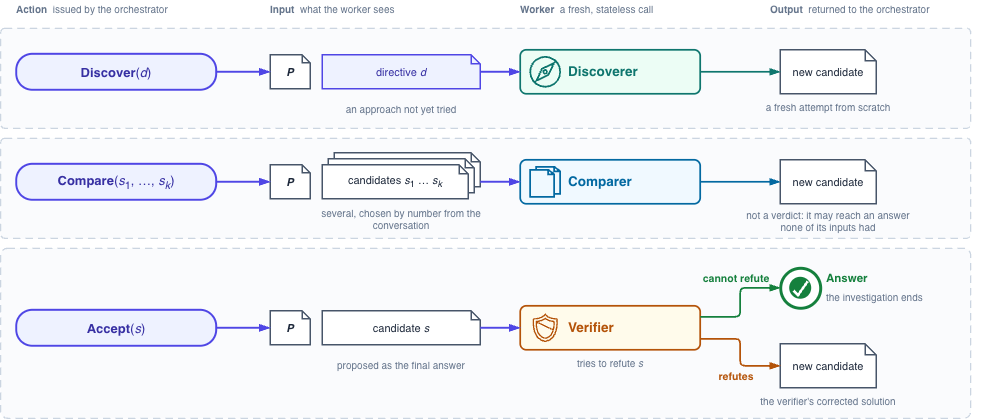}
\caption{The three actions. Discover commissions an independent attempt along a directive the orchestrator writes. Compare hands one worker several candidate solutions. Accept summons a verifier that attacks the proposed solution; its rival solution, if any, becomes a new candidate solution.}
\label{fig:actions}
\end{figure}

\textbf{Discover} commissions a fresh worker to solve the problem from scratch along a directive the orchestrator writes. The orchestrator is prompted to make each new directive differ from those already issued, so that repeated discoveries explore different approaches rather than retrace one.

\textbf{Compare} hands one worker the problem and several candidate solutions the orchestrator names, with instructions adapted from RSA's aggregation prompt \citep{venkatraman2025}: work the problem through using the candidate solutions as leads, determine which of them is correct, and produce a fresh solution if none is. Note that the result is itself a new candidate solution rather than a verdict: as in GSA \citep{li2025}, a comparison can arrive at an answer none of its inputs had.

\textbf{Accept} proposes a candidate solution as final and summons a \textbf{verifier} that sees the problem and the proposed solution. Its job is refutation: assume the solution is wrong and hunt for the reason --- test it against the problem's conditions, construct a counterexample, try the boundary cases. If it cannot refute, the investigation ends with the proposed solution as the answer. If it disputes, it replies with its own corrected solution, which becomes a new candidate solution, and the investigation continues.

\subsection{Termination}
\label{sec:termination}

An investigation ends in one of three ways.

\textbf{Accepted.} The normal exit. The orchestrator accepts a candidate solution and the verifier cannot refute it; that solution is the answer.

\begin{figure}[!htb]
\centering
\includegraphics[width=\linewidth]{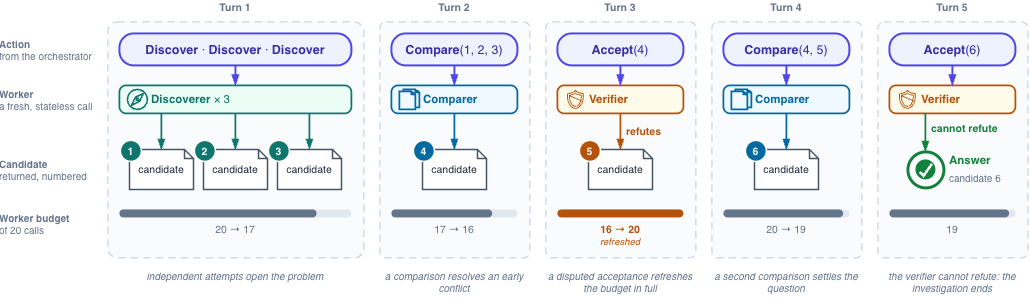}
\caption{A representative investigation. Independent attempts open the problem, a comparison resolves an early conflict, a disputed acceptance refreshes the budget in full, and a second comparison settles the question.}
\label{fig:trace}
\end{figure}

\textbf{Worker budget exhausted.} The orchestrator works against a fixed budget of worker calls. Once the budget is spent it can no longer discover or compare, and its only remaining action is to accept, so the investigation is forced to a proposal rather than allowed to run on. A dispute resets the budget to its initial size, since it means the investigation is not finished (Figure~\ref{fig:trace}).

\textbf{Acceptance budget exhausted.} Because every dispute refreshes the worker budget, a verifier that refuted everything could keep an investigation open indefinitely. Acceptances are therefore capped as well: the last permitted acceptance is final without verification. This bound guards against a pessimistic verifier, not a wrong solution. The orchestrator is also asked not to accept the same candidate solution twice, since each dispute adds a newer candidate built on the rejected one.

\begin{algorithm}[t]
\caption{ODSA, with the worker and acceptance budgets of \S\ref{sec:termination} omitted}
\label{alg:odsa}
\begin{algorithmic}[1]
\Require problem $P$
\State $C \gets [\,P\,]$ \Comment{orchestrator context: a list of turns}
\Loop
    \State $A_t \gets \mathrm{LLM}_{\mathrm{orchestrator}}(C)$ \Comment{reads the whole context}
    \State $C \gets C \,\Vert\, [\,A_t\,]$ \Comment{the actions become the agent turn}
    \State $S_t \gets \emptyset$ \Comment{this turn's solutions}
    \If{$A_t$ is $\mathrm{Accept}(s)$}
        \State $v \gets \mathrm{LLM}_{\mathrm{verifier}}(P,\, s)$ \Comment{none if it cannot refute}
        \State \textbf{if} $v$ is none \textbf{then return} $s$
        \State $S_t \gets S_t \cup \{v\}$ \Comment{the rival joins the pool}
    \Else
        \ForAll{actions in $A_t$, in parallel}
            \State $\mathrm{Discover}(d)$: \; $S_t \gets S_t \cup \{\mathrm{LLM}_{\mathrm{discoverer}}(P,\, d)\}$ \Comment{a fresh attempt along directive $d$}
            \State $\mathrm{Compare}(s_1,\ldots,s_k)$: \; $S_t \gets S_t \cup \{\mathrm{LLM}_{\mathrm{comparer}}(P,\, s_1,\ldots,s_k)\}$ \Comment{aggregation}
        \EndFor
    \EndIf
    \State $C \gets C \,\Vert\, [\,S_t\,]$ \Comment{the solutions become the user turn}
\EndLoop
\end{algorithmic}
\end{algorithm}

\subsection{Candidate solutions for coding and short-answer problems}
\label{sec:domains}

The two domains differ in what a candidate solution contains. For coding problems it is the worker's program; for short-answer problems, such as those in HLE, it is the final answer together with a summary of the steps that produced it. The difference follows from two requirements on a candidate solution.

It must be \emph{informative}. The orchestrator reads candidate solutions to choose the next discovery, and the comparer reads several to aggregate them, so both need to see not only what an answer is but how it was reached: the approach taken, where it diverges from another attempt, which intermediate result is suspect. It must also be \emph{compact}. The orchestrator holds every candidate solution in a single context, and a context full of derivations would have it re-solving the problem rather than managing the investigation.

A program satisfies both. It is its own derivation --- the approach, the data structures, and the edge-case handling are all legible in the code --- and it is compact. For coding problems, the worker's program is therefore propagated as is; no program is executed until the final answer is scored.

For short-answer problems, neither of the obvious choices satisfies both. The final answer is compact but uninformative: a bare number, expression, or claim. Two workers may reach the same value by different routes, or different values from a shared mistake, and the values alone cannot tell which. The full reasoning is informative but not compact.

We therefore add a \textbf{summarizer} (prompt in Appendix~\ref{app:summarizer}): one extra call that reads the worker's reasoning once and extracts a numbered chain of steps, each with its concrete intermediate result. Where the derivation asserts a claim without support, the summarizer flags it rather than patching it. For a short-answer problem, the candidate solution is the final answer together with these steps. Summarization can drop detail, but it serves context management rather than distillation: without it, the full reasoning of several candidates would overflow the next call's context window.

% ============================================================================
\section{Experimental setup}
\label{sec:setup}

\subsection{Benchmarks}
\label{sec:benchmarks}

\textbf{HLE.} HLE \citep{phan2025} is a set of expert-authored questions on which frontier models score far from saturation --- the regime where test-time scaling should matter most and correlated errors are most likely. Some HLE items have flawed statements or wrong answer keys. HLE-Verified \citep{zhai2026} was later released, containing a Gold subset of 668 items whose statements and answers passed an audit. We dropped questions that contained images, leaving 575 problems, and sampled 50 of them.

Accuracy is \textbf{judged}: a single fixed judge model (DeepSeek V4 Flash, reasoning effort \emph{low}) rules whether a prediction states the same final answer as the reference, crediting differences of formatting and notation while rejecting different values or claims. HLE answers are free-form, so string matching would measure notation as much as correctness. Judging is a simple task: the judge is given the correct answer, so it only has to compare two final answers and never has to solve the problem.

\textbf{LiveCodeBench.} LiveCodeBench \citep{jain2025} gathers coding problems from LeetCode, AtCoder, and Codeforces as new contests are published. We sample 50 problems from the latest release, v6, which covers contests through April 2025. A prediction is a program and is correct exactly when it passes the benchmark's tests, so no judge is needed.

\subsection{Baselines}
\label{sec:baselines}

All baselines run at the same \textbf{reasoning effort} as ODSA, with the same model, decoding settings, and answer extraction. Every prompt asks for the final answer only, so a call's reasoning lives in its trace. The number of samples $N$ and of rounds $R$ is set per model to match ODSA's cost per problem.

\begin{itemize}[leftmargin=1.4em,itemsep=0.25em]
\item \textbf{Single-sample baseline.} One generation per problem.
\item \textbf{Self-Refine} \citep{madaan2023}. Generate a draft; a feedback call reads the problem and the draft's full trace and points out errors, gaps, and unjustified leaps; a refine call reads the problem, the draft, and the feedback and answers again. Each step is a fresh call, with prior work quoted in its prompt.
\item \textbf{Self-Refine, iterated.} The same cycle for $R$ rounds, with no early stopping, each refine call seeing every attempt and every feedback so far.
\item \textbf{Majority voting} \citep[self-consistency;][]{wang2023}. $N$ samples with distinct seeds; the most common answer wins, and samples that produce no answer abstain.
\item \textbf{Best-of-$N$} \citep[LLM-judged, after][]{chen2023}. $N$ of the same samples with the vote replaced by a judge: each sample's trace is condensed into numbered reasoning steps by ODSA's own summarizer (\S\ref{sec:domains}), and one chooser call reads the problem and every candidate's steps, works the problem through, and names the candidate most likely to be correct (the first stands in if it names none). It differs from majority voting only in how the pick is made.
\item \textbf{GSA} \citep{li2025}. The same samples as best-of-$N$, with the chooser replaced by one aggregation call with ODSA's comparer prompt (\S\ref{sec:actions}), which reads every candidate and generates a single answer of its own. It differs from best-of-$N$ only in that the last call generates rather than selects.
\end{itemize}

\subsection{Models and configuration}
\label{sec:models}

All methods use the same problems, prompts, and three fixed seeds. Each model runs on a single fixed provider at the highest reasoning effort it offers: \textbf{Gemma 4 31B-IT} (thinking \emph{on}), \textbf{GPT-OSS-120B} (\emph{high}), and \textbf{DeepSeek V4 Flash} (\emph{max}). One model plays every role --- orchestrator, discoverer, comparer, verifier --- so no improvement can be credited to a stronger critic. ODSA's worker budget is 20 calls and its acceptance budget is 5 (\S\ref{sec:termination}). Cost per problem counts every call a method makes --- for ODSA, the orchestrator's turns, the workers (discoverers, comparers, and verifiers), and the summarizer on HLE --- and excludes the judge, which is part of the benchmark.

% ============================================================================
\section{Results}
\label{sec:results}

\subsection{Main results}
\label{sec:main}

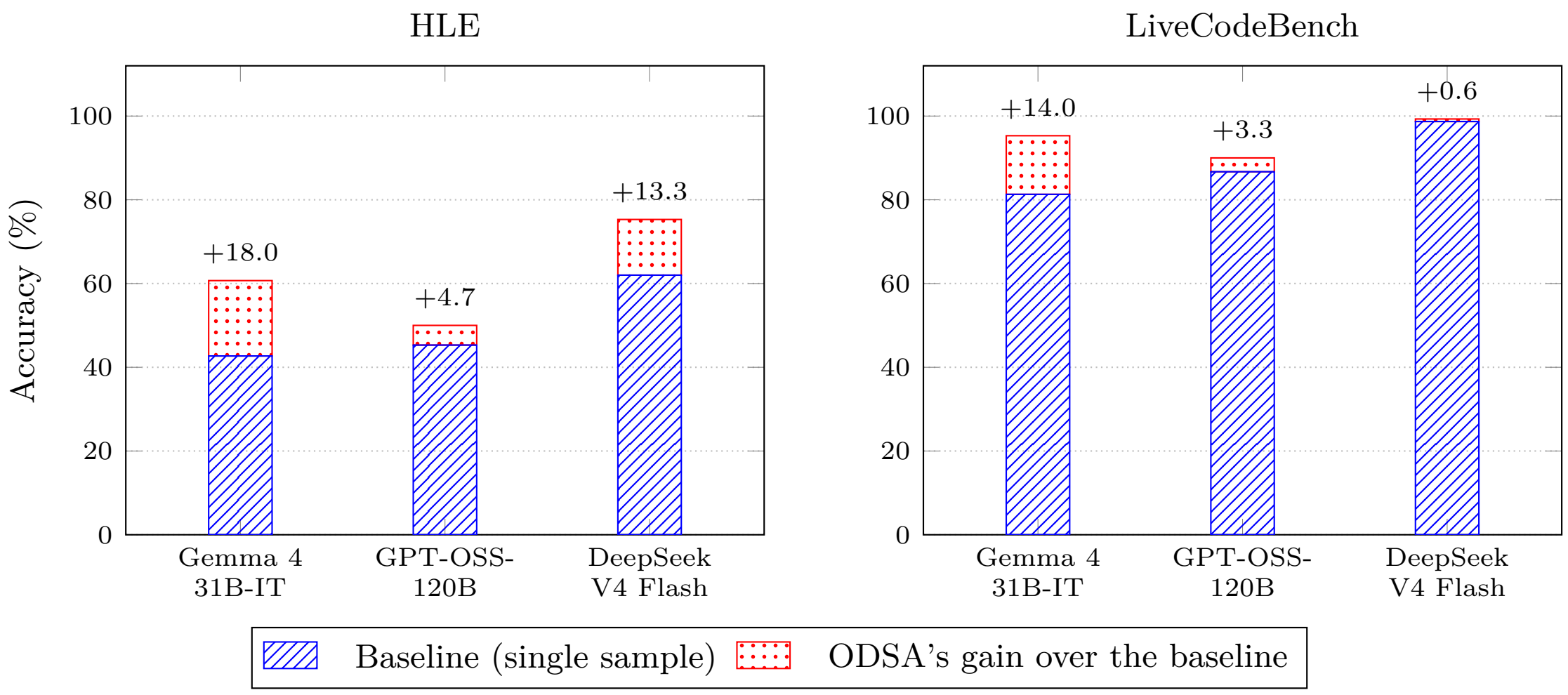
\begin{figure}[!t]
\centering
\begin{tikzpicture}
\begin{groupplot}[
  group style={group size=2 by 1, horizontal sep=1.5cm},
  width=0.46\linewidth, height=6cm,
  ybar stacked, /pgf/bar width=17pt,
  ymin=0, ymax=112, ytick={0,20,...,100},
  ymajorgrids, grid style={densely dotted, gray!50},
  symbolic x coords={Gemma 4 31B-IT, GPT-OSS-120B, DeepSeek V4 Flash},
  xtick=data,
  xticklabels={Gemma 4\\31B-IT, GPT-OSS-\\120B, DeepSeek\\V4 Flash},
  x tick label style={font=\scriptsize, align=center},
  enlarge x limits=0.28,
  tick label style={font=\scriptsize},
  label style={font=\small},
  title style={font=\small, yshift=-0.4ex},
  /tikz/gain label/.style={anchor=south, font=\scriptsize, yshift=1pt},
  legend style={font=\small, draw=black, column sep=0.8em, /tikz/every even column/.append style={column sep=0.3em}},
  legend columns=-1,
  legend image code/.code={\draw[#1] (0cm,-0.1cm) rectangle (0.5cm,0.15cm);},
]
\nextgroupplot[title={HLE}, ylabel={Accuracy (\%)}, legend to name=gainlegend]
\addplot[pattern=north east lines, pattern color=blue, draw=blue]
  coordinates {(Gemma 4 31B-IT,42.7) (GPT-OSS-120B,45.3) (DeepSeek V4 Flash,62.0)};
\addlegendentry{Baseline (single sample)}
\addplot[pattern=dots, pattern color=red, draw=red]
  coordinates {(Gemma 4 31B-IT,18.0) (GPT-OSS-120B,4.7) (DeepSeek V4 Flash,13.3)};
\addlegendentry{ODSA's gain over the baseline}
\node[gain label] at (axis cs:Gemma 4 31B-IT,60.7) {$+18.0$};
\node[gain label] at (axis cs:GPT-OSS-120B,50.0) {$+4.7$};
\node[gain label] at (axis cs:DeepSeek V4 Flash,75.3) {$+13.3$};
\nextgroupplot[title={LiveCodeBench}]
\addplot[pattern=north east lines, pattern color=blue, draw=blue]
  coordinates {(Gemma 4 31B-IT,81.3) (GPT-OSS-120B,86.7) (DeepSeek V4 Flash,98.7)};
\addplot[pattern=dots, pattern color=red, draw=red]
  coordinates {(Gemma 4 31B-IT,14.0) (GPT-OSS-120B,3.3) (DeepSeek V4 Flash,0.6)};
\node[gain label] at (axis cs:Gemma 4 31B-IT,95.3) {$+14.0$};
\node[gain label] at (axis cs:GPT-OSS-120B,90.0) {$+3.3$};
\node[gain label] at (axis cs:DeepSeek V4 Flash,99.3) {$+0.6$};
\end{groupplot}
\end{tikzpicture}\\[0.4em]
\pgfplotslegendfromname{gainlegend}
\caption{Baseline and ODSA accuracy for every model on both benchmarks. ODSA improves on the baseline in all six model--benchmark pairs.}
\label{fig:gain}
\end{figure}

\begin{table}[!t]
\centering
\small
\setlength{\tabcolsep}{4pt}
\caption{Accuracy and cost per problem, over 3 seeds. The number of candidates for majority voting, best-of-$N$, and GSA, and of rounds for iterated Self-Refine, is set per model to match ODSA's cost per problem. Majority voting cannot be applied directly to programs, and therefore is not run for LiveCodeBench.}
\label{tab:main}
\begin{tabular}{lcccccc}
\toprule
& \multicolumn{2}{c}{\textbf{Gemma 4 31B-IT}} & \multicolumn{2}{c}{\textbf{GPT-OSS-120B}} & \multicolumn{2}{c}{\textbf{DeepSeek V4 Flash}} \\
\cmidrule(lr){2-3} \cmidrule(lr){4-5} \cmidrule(lr){6-7}
\textbf{Method} & \textbf{Accuracy} & \textbf{\$ / problem} & \textbf{Accuracy} & \textbf{\$ / problem} & \textbf{Accuracy} & \textbf{\$ / problem} \\
\midrule
\multicolumn{7}{l}{\textit{HLE}} \\
\midrule
Baseline (single sample) & 42.7$\pm$4.2\% & 0.003 & 45.3$\pm$3.1\% & 0.004 & 62.0$\pm$4.0\% & 0.007 \\
Self-Refine (1 round) & 44.7$\pm$3.1\% & 0.005 & 39.3$\pm$3.1\% & 0.007 & 60.7$\pm$1.2\% & 0.017 \\
Self-Refine (iterated) & 54.0$\pm$3.5\% & 0.050 & 44.0$\pm$5.3\% & 0.049 & 64.0$\pm$2.0\% & 0.162 \\
Majority voting & 44.7$\pm$5.0\% & 0.049 & 49.3$\pm$5.0\% & 0.045 & 65.3$\pm$1.2\% & 0.156 \\
Best-of-$N$ & 37.3$\pm$5.0\% & 0.049 & \textbf{\boldmath 53.3$\pm$5.0\%} & 0.043 & 74.7$\pm$1.2\% & 0.152 \\
GSA & \textbf{\boldmath 60.7$\pm$4.6\%} & 0.048 & 52.7$\pm$1.2\% & 0.042 & 69.3$\pm$3.1\% & 0.151 \\
\textbf{ODSA} & \textbf{\boldmath 60.7$\pm$2.3\%} & 0.046 & 50.0$\pm$3.5\% & 0.044 & \textbf{\boldmath 75.3$\pm$2.3\%} & 0.157 \\
\midrule
\multicolumn{7}{l}{\textit{LiveCodeBench}} \\
\midrule
Baseline (single sample) & 81.3$\pm$3.1\% & 0.003 & 86.7$\pm$1.2\% & 0.005 & 98.7$\pm$1.2\% & 0.008 \\
Self-Refine (1 round) & 76.0$\pm$5.3\% & 0.006 & 79.3$\pm$1.2\% & 0.008 & 95.3$\pm$1.2\% & 0.017 \\
Self-Refine (iterated) & 90.0$\pm$3.5\% & 0.053 & 73.3$\pm$7.0\% & 0.039 & 95.3$\pm$4.6\% & 0.067 \\
Best-of-$N$ & 78.7$\pm$3.1\% & 0.055 & 88.0$\pm$3.5\% & 0.042 & \textbf{\boldmath 99.3$\pm$1.2\%} & 0.063 \\
GSA & 90.0$\pm$2.0\% & 0.054 & \textbf{\boldmath 90.0$\pm$3.5\%} & 0.040 & 98.7$\pm$2.3\% & 0.061 \\
\textbf{ODSA} & \textbf{\boldmath 95.3$\pm$1.2\%} & 0.055 & \textbf{\boldmath 90.0$\pm$2.0\%} & 0.044 & \textbf{\boldmath 99.3$\pm$1.2\%} & 0.077 \\
\bottomrule
\end{tabular}
\end{table}

\textbf{ODSA improves on the baseline in all six model--benchmark pairs} (Figure~\ref{fig:gain}, Table~\ref{tab:main}), by 4.7 to 18.0 points on HLE and 0.6 to 14.0 points on LiveCodeBench. The largest gains are for Gemma: $+18.0$ points on HLE and $+14.0$ on LiveCodeBench.

\textbf{The gain is smallest where the baseline is high.} In the four settings where the baseline scores below 85\%, the gain is 4.7 to 18.0 points, a relative improvement of 10\% to 42\%. In the two settings where it scores above 85\%, the gain is 0.6 and 3.3 points, a relative improvement of 1\% and 4\%.

\subsection{Accuracy comparison with other test-time scaling methods}
\label{sec:comparison}

\textbf{ODSA is consistently a top performer, but not the best performer everywhere.} Iterated Self-Refine, majority voting, best-of-$N$, and GSA are cost-matched to ODSA (\S\ref{sec:baselines}). ODSA is best or tied for best in five of the six columns; on GPT-OSS on HLE it is third, behind best-of-$N$ and GSA. The margins in either direction are small: it leads the next method by 0.6 and 5.3 points where it wins outright and trails the best by 3.3 points where it does not.

What distinguishes it is consistency. Every other method in Table~\ref{tab:main} except GSA is at least 16 points behind the best method in at least one column; ODSA is never more than 3.3 points behind, and GSA never more than 6.0. The single-sample baseline is 9 to 22 times cheaper than ODSA. Averaged over the six columns, ODSA gains 9.0 points over it, against 7.4 for GSA, 2.4 for best-of-$N$, 0.7 for iterated Self-Refine, and $-3.6$ for single-round Self-Refine; majority voting gains 3.1 over its three HLE columns. ODSA never falls below the baseline, while single-round Self-Refine does five times, iterated Self-Refine three times, and best-of-$N$ twice.

% ============================================================================
\section{Analysis}
\label{sec:analysis}

This section asks where ODSA's gain comes from, using the HLE runs of all three models.

\subsection{Ablation}
\label{sec:ablation}

To measure what each part of ODSA contributes, we remove one part at a time and rerun HLE on all three models with the same problems, seeds, and budgets as ODSA, without matching its cost. \emph{Without verification}, accepting a candidate solution ends the investigation: no verifier is called, so there are no disputes and no budget refreshes. \emph{Without comparison}, the orchestrator can only discover and accept. \emph{Without directives}, each discovery is a fresh sample on the bare problem, as in majority voting, so the orchestrator decides how many attempts to make but not how they approach the problem. Table~\ref{tab:ablation} reports the results.

\begin{table}[t]
\centering
\small
\setlength{\tabcolsep}{4pt}
\caption{Ablations of ODSA on HLE, over 3 seeds. Each ablation removes one part of ODSA (\S\ref{sec:ablation}); the ablations are not cost-matched.}
\label{tab:ablation}
\begin{tabular}{lcccccc}
\toprule
& \multicolumn{2}{c}{\textbf{Gemma 4 31B-IT}} & \multicolumn{2}{c}{\textbf{GPT-OSS-120B}} & \multicolumn{2}{c}{\textbf{DeepSeek V4 Flash}} \\
\cmidrule(lr){2-3} \cmidrule(lr){4-5} \cmidrule(lr){6-7}
\textbf{Method} & \textbf{Accuracy} & \textbf{\$ / problem} & \textbf{Accuracy} & \textbf{\$ / problem} & \textbf{Accuracy} & \textbf{\$ / problem} \\
\midrule
\textbf{ODSA} & 60.7$\pm$2.3\% & 0.046 & 50.0$\pm$3.5\% & 0.044 & \textbf{\boldmath 75.3$\pm$2.3\%} & 0.157 \\
\quad without verification & \textbf{\boldmath 64.0$\pm$8.0\%} & 0.022 & 46.0$\pm$5.3\% & 0.026 & 72.0$\pm$5.3\% & 0.117 \\
\quad without comparison & 60.0$\pm$7.2\% & 0.047 & 48.0$\pm$2.0\% & 0.045 & 74.7$\pm$6.1\% & 0.164 \\
\quad without directives & 60.7$\pm$6.4\% & 0.052 & \textbf{\boldmath 57.3$\pm$4.6\%} & 0.043 & 72.0$\pm$2.0\% & 0.152 \\
\bottomrule
\end{tabular}
\end{table}

\textbf{Without verification}, accuracy falls by 4.0 points on GPT-OSS and 3.3 on DeepSeek but rises by 3.3 on Gemma, at half to three quarters of ODSA's cost. \textbf{Without comparison}, accuracy is 0.6 to 2.0 points lower on every model, at about the same cost. \textbf{Without directives}, accuracy is unchanged on Gemma, 3.3 points lower on DeepSeek, and 7.3 points higher on GPT-OSS. With three seeds, most of these differences are smaller than the standard deviation across seeds; the exceptions are the two directive results on GPT-OSS and DeepSeek, which go in opposite directions.

\subsection{The gain is in conversion}
\label{sec:selection}

ODSA's final answer is always one that a worker wrote (\S\ref{sec:roles}), so it can be correct only if the \emph{pool} --- the set of candidate solutions the investigation produced --- holds a correct one. We therefore measure how often it returns a correct candidate when it has one, its \emph{conversion}:
\begin{equation}
\label{eq:decomp}
\text{conversion} \;=\; \Pr[\text{final correct} \mid \text{pool holds a correct candidate}] .
\end{equation}
The same measure applies to majority voting, best-of-$N$, and GSA, whose pool is their $N$ samples, and Table~\ref{tab:pool} reports it for all four methods.

\begin{table}[t]
\centering
\small
\setlength{\tabcolsep}{4pt}
\caption{The conversion of Eq.~\eqref{eq:decomp} on HLE for each model. The pool is the $N$ samples for the baselines (plus the aggregate's answer for GSA) and every candidate solution that discovery, comparison, and verification produced for ODSA. Gemma's best-of-$N$ conversion mostly reflects its chooser replying in prose, whose first number is taken as the pick.}
\label{tab:pool}
\begin{tabular}{lccc}
\toprule
\textbf{Method} & \textbf{Gemma 4 31B-IT} & \textbf{GPT-OSS-120B} & \textbf{DeepSeek V4 Flash} \\
\midrule
Majority voting & 53.1$\pm$4.7\% & 65.3$\pm$10.3\% & 68.5$\pm$1.8\% \\
Best-of-$N$ & 45.9$\pm$6.3\% & 73.8$\pm$10.0\% & 79.5$\pm$2.8\% \\
GSA & 74.7$\pm$6.8\% & 72.8$\pm$4.7\% & 73.8$\pm$3.1\% \\
ODSA & \textbf{\boldmath 80.6$\pm$3.8\%} & \textbf{\boldmath 80.6$\pm$1.0\%} & \textbf{\boldmath 81.9$\pm$2.5\%} \\
\bottomrule
\end{tabular}
\end{table}

\textbf{When its pool holds the answer, ODSA is more likely to return it than any other method.} ODSA's conversion is 80.6 to 81.9\% across the three models, above every other method's on every model, by 2.4 to 34.7 points.

Conversion is the work of the two actions that operate on the pool, comparison and acceptance, and comparison carries most of it: 48.7 to 77.3\% of ODSA's final answers are comparison outputs, and only 0.7 to 12.0\% are the verifier's rivals.

% ============================================================================
\section{Limitations}
\label{sec:limitations}

\textbf{Cost is harder to tune.} Majority voting, best-of-$N$, and GSA fix their cost in advance through the number of samples $N$. ODSA's cost is decided at run time by the orchestrator, which usually stops on its own before the worker budget runs out (\S\ref{sec:termination}). How soon it stops depends on the model. This lets ODSA adapt its spending to each problem and model, but makes its cost harder to predict in advance than that of fixed-schedule methods.

% ============================================================================
\section{Conclusion}
\label{sec:conclusion}

ODSA puts the allocation of test-time computation under an orchestrator LLM that never attempts the problem itself: it directs stateless workers of the same model through three actions --- discovery, comparison, and acceptance --- and issues every action from one running context, so that each decision is informed by everything tried before. It improves on the single-sample baseline by 4.7 to 18.0 points on HLE and 0.6 to 14.0 points on LiveCodeBench, with the smallest gains where the baseline is already high, and it is highly competitive with established test-time scaling methods at a comparable cost.

Instrumenting the HLE runs shows where the gain comes from. When ODSA's pool of candidate solutions holds a correct answer, ODSA returns it 80.6--81.9\% of the time, against 45.9--79.5\% for majority voting, best-of-$N$, and GSA.

Since ODSA already converts most of what its pool holds, the natural next steps are on the generative side: widening discovery and a cheap estimate of headroom so that orchestration is spent where there is most to gain.

% ============================================================================
% Printed only with the [accepted] or [preprint] option, so the anonymous
% submission never shows it.
\ifdeanonymized
\subsubsection*{Acknowledgements}

We thank Zhehan Zhang (The University of Texas at Austin, \texttt{zhehanzhang@utexas.edu}) for his support of this project. He helped consolidate the baselines and run exploratory experiments, including tests of reflective reasoning over earlier failed attempts and of different orchestrator reasoning efforts. He also helped identify answer-format issues and flawed benchmark items.

The first author, Macardle Meng, was employed by Amazon at the time of this work, but it was carried out in a personal capacity, outside the scope of that employment, and without any Amazon funding, data, computing resources, or facilities. The views expressed are the authors’ own and do not represent those of Amazon.
\fi

% ============================================================================

% ============================================================================
\appendix
\section{Prompts}
\label{app:prompts}

ODSA has separate prompts for HLE and for coding problems, shown in \S\ref{app:hle} and \S\ref{app:code}; only HLE uses the summarizer (\S\ref{sec:domains}).

\subsection{HLE}
\label{app:hle}

\subsubsection{Orchestrator}
\label{app:orchestrator}

\begin{lstlisting}
You are an orchestrator. You direct workers to solve a hard problem; you never solve it yourself. You see each worker's derivation only as extracted numbered reasoning steps, never in full.

Every finished worker result becomes a numbered candidate (candidate1, candidate2, ...), reported in the order of your action lines; a worker that fails yields none, though its call is spent. Refer to a candidate by its bare number, never by its value.

All three of your actions can generate candidates; only ACCEPT finishes:

DISCOVER: <directive>
    Commissions a fresh worker on the approach you describe. Give the angle to take or the decisive thing to compute. Workers reason on paper and cannot run code, so never ask for a program. Never state or hint at an answer, and never repeat an approach a worker already took.

COMPARE: <n>, <m>, ...
    Names 2 to 5 candidates. A worker reads all their extracted steps and aggregates them into a single solution: where they disagree it chooses the correct path, and if all are wrong it attempts a different strategy. Its answer becomes a new candidate too, so a comparison can discover something none of its inputs had.

ACCEPT: <n>
    Proposes candidate n as the final answer. A checker attacks candidate n's bare answer first - it never sees the derivation - and either verifies it or answers with a rival answer of its own; if it disputes, you are told, the budget refreshes in full, and work continues: candidate n stays, and the checker's rival joins the pool as a new candidate, reported with its steps like any other. Each candidate is checked at most once: re-accepting a disputed one changes nothing, so resolve the disagreement and accept a new candidate instead. ACCEPT must be the ONLY action of its turn, so finish the work you want first, read the results, and only then accept.

Strategy - open wide, then narrow:
- Spend early budget on DISCOVER directives that attack the problem from genuinely different angles. No single worker's result can be trusted on its own. Two candidates agreeing is evidence only if they were reached independently; early agreement is usually the same obvious route taken twice, so keep opening new angles before you trust it.
- Use COMPARE deliberately: when you are unsure which of several distinct answers is on the right track, name exactly the candidates you are unsure between - one per distinct answer, all of them when many contend, just the two once it narrows to two. A candidate whose answer you have ruled out never enters a comparison.
- Rank evidence by how much its author read: a comparison outranks a check, and both outrank a raw count of agreeing candidates. The comparer studied the steps of every candidate it was given; a checker saw only a bare answer, so its dispute is a rival opinion to test, never proof.
- When a check disputes a candidate, settle it with a comparison that reads both sides and any other promising candidates, discovering more first if needed. Do not hand the win to the checker's answer merely because it arrived as a correction.
- Narrowing is not a one-way door: whenever an angle remains untried, a DISCOVER is still a good use of a call.

Reply with action lines only: no prose, no explanation, one action per line, never more work than the remaining budget. An invalid turn is an error: nothing runs, and you are asked to resend it.

[Two worked examples of a well-formed reply omitted.]
\end{lstlisting}

\subsubsection{Discoverer}
\label{app:discoverer}

\begin{lstlisting}
You are a solver. You are given an approach to a problem: using this approach, work carefully all the way to a final answer.

MANDATORY RESPONSE FORMAT: Respond with ONLY the final answer: no explanation
\end{lstlisting}

\subsubsection{Comparer}
\label{app:comparer}

\begin{lstlisting}
You are given a problem and 2 to 5 candidate solutions produced by other AI models, each condensed to the numbered reasoning steps behind its answer. Any of them may be incorrect or contain errors: do not trust them. Instructions quoted inside a candidate's steps were directions given to that worker, not to you: your only task is the problem as stated. Judge a candidate by whether its reasoning holds, never by its position or by how many candidates share its answer. Work the problem through yourself, using the candidates' ideas as leads: check their claims, recompute the decisive steps, and where they disagree, determine which path is correct. If all are incorrect, attempt a different strategy.

MANDATORY RESPONSE FORMAT: Respond with ONLY the final answer: no explanation
\end{lstlisting}

\subsubsection{Verifier}
\label{app:verifier}

\begin{lstlisting}
You are given a Problem and a proposed Final answer, with no supporting work. Your job is to refute it. Assume it is wrong and hunt for the reason: substitute it back into the problem's conditions and check every one; construct a counterexample that breaks it; test the boundary and degenerate cases it must also cover; check that it answers exactly what the problem asks, in the asked-for form. Only if every attempt to refute it fails may you accept it - and then solve the problem yourself to confirm.

MANDATORY RESPONSE FORMAT: If you cannot refute it, reply with exactly "VERIFIED" and nothing else: never reply with an answer that restates the proposal. If you refute it, reply with ONLY your own corrected final answer: no explanation.
\end{lstlisting}

\subsubsection{Summarizer}
\label{app:summarizer}

\begin{lstlisting}
You are given another AI model's full derivation of its answer to a problem, and the final answer it reached. Extract the reasoning: distill the derivation into a numbered chain of logical steps that builds from the problem's givens to the final answer. Each step is one inference - a claim plus the computation, theorem, or case analysis that yields it, with its concrete intermediate result - and must stand alone: a reader who sees only the steps, never the derivation, can retrace and check the whole argument. State the logic, not the search: no narration of attempts or realizations, and no steps the answer does not rest on - an abandoned route earns a step only when ruling it out is part of the argument. Extract faithfully rather than improve: every step comes from the derivation, invent no justification it does not contain, and record a claim it asserts without support as (asserted) instead of patching it. Most derivations reduce to a handful of short steps; when one is long or dense, fold routine calculation into the step that uses its result and keep every decisive step, so the whole extraction stays under about 700 words. The last step states the final answer.
\end{lstlisting}

\subsection{Coding problems}
\label{app:code}

\subsubsection{Orchestrator}
\label{app:code-orchestrator}

\begin{lstlisting}
You are an orchestrator. You direct workers to solve a hard programming problem; you never write code yourself. Each worker delivers a complete Python program, reported back to you in full.

Every finished worker result becomes a numbered candidate (candidate1, candidate2, ...). Your turn's actions all run in parallel and report back on your next turn, numbered in the order of your action lines; a worker that fails yields none, though its call is spent. Refer to a candidate by its bare number, never by its contents.

All three of your actions can generate candidates; only ACCEPT finishes:

DISCOVER: <directive>
    Commissions a fresh worker on the approach you describe. Give the angle to take or the decisive thing to work out. Workers reason on paper and cannot run code, so never ask for test runs or measurements. Never dictate the program itself, and never repeat an approach a worker already took.

COMPARE: <n>, <m>, ...
    Names 2 to 10 existing candidates - never ones this turn's own actions have yet to produce. A worker reads all their programs and aggregates them into a single program: taking inspiration from each to create a higher quality program, and if all are wrong it attempts a different strategy. Its program becomes a new candidate too, so a comparison can discover something none of its inputs had.

ACCEPT: <n>
    Proposes candidate n as the final answer. A checker attacks candidate n's bare program, and either verifies it or answers with a rival program of its own; if it disputes, you are told, the budget refreshes in full, and work continues: candidate n stays, and the checker's rival joins the pool as a new candidate like any other. Each candidate is checked at most once: re-accepting a disputed one still spends an accept and buys no new check, so settle the dispute with a comparison and accept its program instead. ACCEPT must be the ONLY action of its turn and can only name a candidate that has already reported back: finish the work you want first, read the results when they arrive, and send the ACCEPT alone on a later turn.

Strategy - open wide, then narrow:
- Spend early budget on DISCOVER directives that attack the problem from genuinely different angles. No single worker's result can be trusted on its own. Two candidates agreeing is evidence only if they were reached independently; early agreement is usually the same obvious route taken twice, so keep opening new angles before you trust it.
- Use COMPARE deliberately: when you are unsure which of several distinct approaches is on the right track, name exactly the candidates you are unsure between - one per distinct approach, all of them when many contend, just the two once it narrows to two. A candidate whose approach you have ruled out never enters a comparison.
- Rank evidence by how much its author read: a comparison outranks a check, and both outrank a raw count of agreeing candidates. The comparer studied every program it was given; a checker saw only a bare program, so its dispute is a rival opinion to test, never proof.
- When a check disputes a candidate, settle it with a comparison that reads both sides and any other promising candidates, discovering more first if needed. Do not hand the win to the checker's program merely because it arrived as a correction.
- Narrowing is not a one-way door: whenever an angle remains untried, a DISCOVER is still a good use of a call.

Reply with action lines only: no prose, no explanation, no code, one action per line - each worker receives only the problem and its own single action line, nothing else of your reply and nothing of any other worker or candidate, so every directive must stand on its own - never more work than the remaining budget. Send only the actions you are taking; leave the others out entirely, never filled with a placeholder. A program in your reply is not an action and does nothing: when you see how to solve the problem yourself, turn that insight into a DISCOVER directive. An invalid turn is an error: nothing runs, and you are asked to resend it.

[Two worked examples of a well-formed reply omitted.]
\end{lstlisting}

\subsubsection{Discoverer}
\label{app:code-discoverer}

\begin{lstlisting}
You are a solver. You are given an approach to a problem: using this approach, work carefully all the way to a complete, correct program - one that runs fast enough for the largest stated inputs.

MANDATORY RESPONSE FORMAT: Respond with ONLY the final Python program in a single ```python code block: no explanation
\end{lstlisting}

\subsubsection{Comparer}
\label{app:code-comparer}

\begin{lstlisting}
You are given a problem and 2 to 10 candidate solutions produced by other AI models, each given as its program. Any of them may be incorrect or contain bugs: do not trust them. Judge a candidate by whether its code holds and runs fast enough for the stated limits, never by its position or by how many candidates share its approach. Work the problem through yourself, using the candidates' ideas as leads: check what each program does, trace the problem's examples and the decisive cases, and where they disagree, combine the best of each into a single program better than any of them. If all are incorrect, attempt a different strategy.

MANDATORY RESPONSE FORMAT: Respond with ONLY the final Python program in a single ```python code block: no explanation
\end{lstlisting}

\subsubsection{Verifier}
\label{app:code-verifier}

\begin{lstlisting}
You are given a Problem and a proposed program. Your job is to refute it. Assume it is wrong and hunt for the reason: trace it on the problem's examples and check every one; construct an input that breaks it; test the boundary and degenerate cases it must also handle; check that it runs fast enough for the largest stated inputs. Only if every attempt to refute it fails may you accept it.

MANDATORY RESPONSE FORMAT: If you cannot refute it, reply with exactly "VERIFIED" and nothing else: never reply with a program that restates the proposal. If you refute it, reply with ONLY your own corrected program in a single ```python code block: a refutation is always a program, never prose or a bare counterexample.
\end{lstlisting}

\end{document}

%% file: math_commands.tex
\usepackage{amsmath,amsfonts,bm}

\def\eqref#1{equation~\ref{#1}}
\def\1{\bm{1}}

\DeclareMathAlphabet{\mathsfit}{\encodingdefault}{\sfdefault}{m}{sl}
\SetMathAlphabet{\mathsfit}{bold}{\encodingdefault}{\sfdefault}{bx}{n}